\documentclass[conference]{IEEEtran}
\usepackage{times}

\usepackage{graphics} %
\usepackage{epsfig} %
\usepackage{mathptmx} %
\usepackage{times} %
\usepackage{amsmath} %
\usepackage{amssymb}  %

\usepackage{booktabs}
\usepackage[table]{xcolor}
\usepackage{vcell}
\usepackage{subcaption}
\usepackage{graphicx}
\usepackage{amsmath}
\usepackage{float}
\usepackage{amssymb}
\usepackage{tikz}
\usepackage{xspace}
\usepackage{bm}
\usepackage{wrapfig}
\usepackage{multirow}
\usepackage{duckuments}
\usepackage{graphicx}
\usepackage{pifont} %
\usepackage{listings}

\newcommand{\smallsec}[1]{\noindent {\bf #1.}}
\newcommand{\method}{RVT-2\xspace}

\newcommand{\tb}[1]{\textbf{#1}}
\newcommand{\yes}{\color{blue}{\ding{51}}}
\newcommand{\no}{\color{red}{\ding{55}}}

\newcommand{\rpmh}{\huge \raisebox{.2ex}{$\scriptstyle\pm~$}}

\newcommand{\todo}[1]{\textcolor{black}{{ }}}

\newcommand{\bczcnn}{Image-BC (CNN)\xspace}
\newcommand{\bczvit}{Image-BC (ViT)\xspace}
\newcommand{\unet}{C2F-ARM-BC\xspace}
\newcommand{\peract}{PerAct\xspace}
\newcommand{\purl}{\url{https://robotic-view-transformer-2.github.io/}}

\usepackage[numbers]{natbib}
\usepackage{multicol}
\usepackage[bookmarks=true]{hyperref}

\begin{document}

\title{\method: Learning Precise Manipulation \\ from Few Demonstrations}

\makeatletter
\g@addto@macro\@maketitle{
  \begin{figure}[H]
  \setlength{\linewidth}{\textwidth}
  \setlength{\hsize}{\textwidth}
  \centering
  \resizebox{0.98\textwidth}{!}{
    \includegraphics[width=\textwidth]{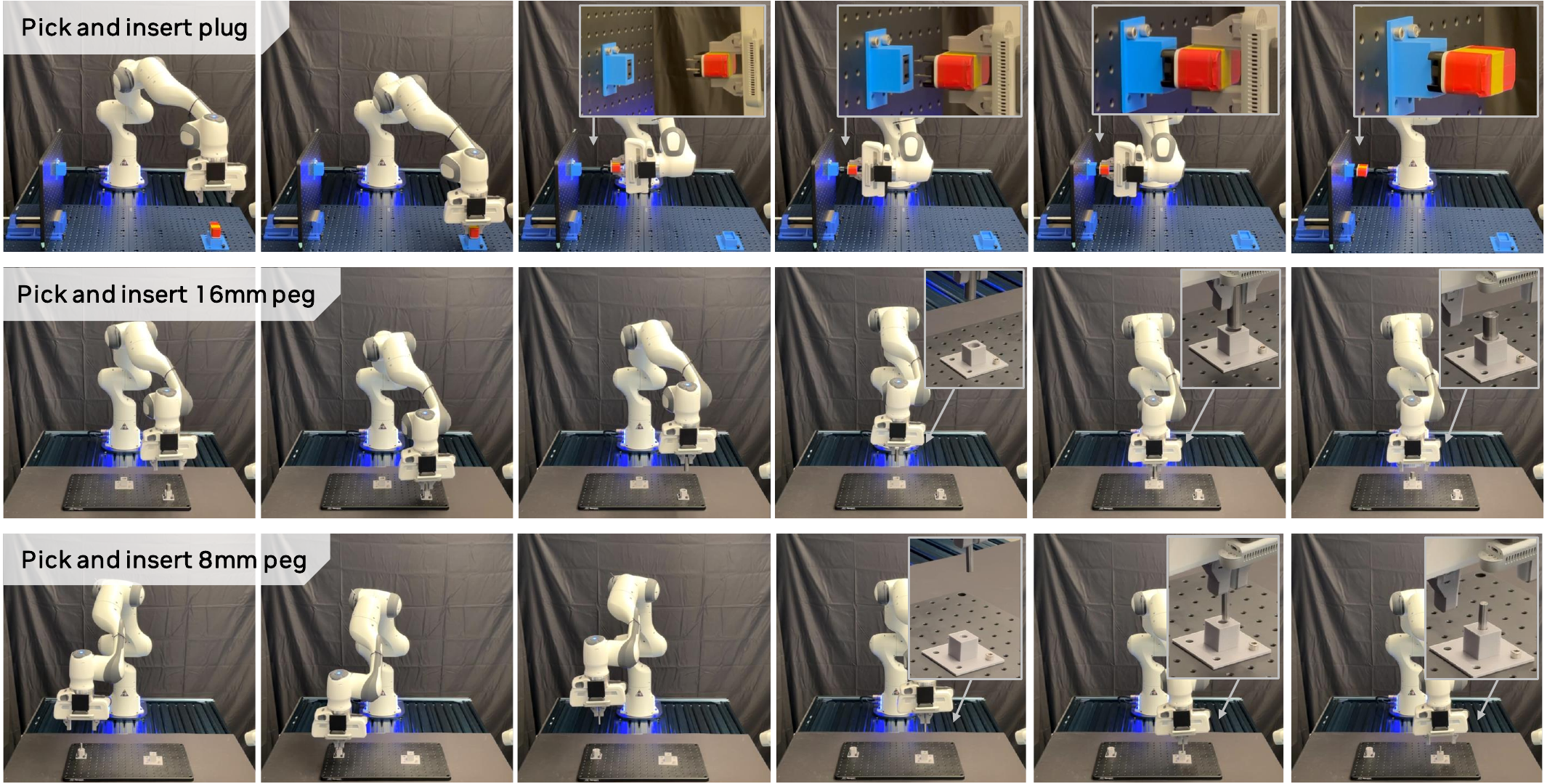}
  }
  \caption{\textbf{\method performing high precision tasks.} Given a language instruction, a single \method model can perform multiple 3D manipulation tasks, including ones requiring millimeter-level precision like \textit{inserting peg in hole} and \textit{inserting plug in socket}. \method is trained with $\sim$10 demonstrations per task and uses only a single third-person RGB-D camera.}
  \label{fig:teaser}
  \vspace{-3mm}
  \end{figure}
}
\makeatother

\author{Ankit Goyal, Valts Blukis, Jie Xu, Yijie Guo, Yu-Wei Chao, Dieter Fox \\ NVIDIA}

\maketitle
\begin{abstract}
In this work, we study how to build a robotic system that can solve multiple 3D manipulation tasks given language instructions. To be useful in industrial and household domains, such a system should be capable of learning new tasks with few demonstrations and solving them precisely. Prior works, like PerAct~\cite{shridhar:corl2022} and RVT~\cite{goyal:corl2023}, have studied this problem, however, they often struggle with tasks requiring high precision. We study how to make them more effective, precise, and fast. Using a combination of architectural and system-level improvements, we propose \method, a multitask 3D manipulation model that is 6X faster in training and 2X faster in inference than its predecessor RVT. \method achieves a new state-of-the-art on RLBench~\cite{james:ral2020}, improving the success rate from 65\% to 82\%. \method is also effective in the real world, where it can learn tasks requiring high precision, like picking up and inserting plugs, with just 10 demonstrations. Visual results, code, and trained model are provided at: \purl.

\end{abstract}

\IEEEpeerreviewmaketitle

\setcounter{figure}{1}

\section{Introduction}
\label{sec:intro}
One of the holy grails of robot learning is building general-purpose robotic systems that can solve multiple tasks and generalize to unseen environment configurations. To be useful, such systems should be capable of precise manipulation and only need a few demonstrations of a new task. For example, in an industrial manufacturing setting, we can expect a person to demonstrate a high-precision task like peg insertion to a robot just a few times, after which the robot should start doing that task independently. Similar examples can be found in other domains like household and retail. In this work, we study the problem of building a manipulation system that can solve various tasks precisely, given just a few demonstrations. The systems should have three key characteristics: (1) handle multiple tasks, (2) require only a few demonstrations, and (3) solve tasks with high precision.

Prior work has made significant progress towards this goal. 
Starting with works like Transporter Networks~\cite{zeng:corl2020} and IFOR~\cite{goyal:cvpr2022} that studied planar pick-and-place tasks, recent works have gone beyond the 2D plane and studied manipulation in 3D with a few examples~\cite{james:cvpr2022}. Some notable methods are PerAct~\cite{shridhar:corl2022} and RVT~\cite{goyal:corl2023}. Given a language instruction, PerAct~\cite{shridhar:corl2022} adopted a multi-task transformer model for 3D manipulation by predicting the next keyframe pose. Even though PerAct achieved impressive performance, it uses a voxel-based representation for the scene, limiting its scalability. RVT~\cite{goyal:corl2023} addressed the limitations of PerAct by proposing a novel multi-view representation for encoding the scene. The multi-view representation has various advantages, including faster training speed, faster inference, and better task performance. Compared to PerAct, RVT demonstrated a 36X faster training speed and improved the performance from $48\%$ to $63\%$ on 18 tasks in RLBench~\cite{james:ral2020}.

We were motivated by the question of what prevents RVT from achieving even higher performance. Upon careful analysis, we find that RVT struggles with tasks requiring high precision, like \textit{screwing bulb} or \textit{inserting a peg}. During our analysis, we also identified several opportunities to further improve the training and inference speed of the system. Through our architectural and system-level improvements, we were able to boost both the speed and efficacy of RVT. We thereby present \method, which improves RVT on the training speed by 6X (from 2.4M samples per day to 16M samples per day), inference speed by 2X (from 11.6 fps to 20.6 fps), and task success rate by 15 points (from 62.9 to 77.6) on the RLBench benchmark, achieving state-of-the-art results. 

We also find that a single \method model is able to solve multiple tasks in the real world with as few as 10 demonstrations. Specifically, \method can perform tasks requiring millimeter-level precision, like \textit{inserting a peg in a hole} and \textit{inserting a plug in a socket}, while only using a single third-person camera. To the best of our knowledge, this is the first time a vision-based policy trained with a few examples has been tested to work on such high-precision tasks. 

Overall, the gains in \method were achieved by a combination of architectural and system-level improvements. For the architectural improvement, we introduce three main design innovations. First, we equip \method with a multi-stage inference pipeline that allows the network to zoom into the region of interest and predict more precise end-effector poses. Further, to save GPU memory during training and to improve speed, we adopt a convex upsampling technique. Lastly, we improve end-effector rotation prediction by utilizing location-conditioned features instead of just the global features as done in RVT.

For the system-level optimizations, we create a custom virtual image renderer to replace the generic renderer used in RVT (PyTorch3D~\cite{ravi:arxiv2020}). With this custom accelerated rendering library, we improve the speed and reduce the memory usage of RVT in both training and inference. We also investigate and incorporate cutting-edge practices in training transformer models, including fast, optimized optimizers and mixed-precision training. While each of these changes in itself is not novel and has appeared in some form in prior works,  our contribution lies in building a precise 3D manipulation system by incorporating these changes successfully.  

To summarize, we push the frontiers of 3D manipulation with few-shot demonstrations. We achieve significant improvements and demonstrate superior real-world performance. We also provide a careful analysis ablating and quantifying the improvements sourced from different factors. Our code is available at \purl.

\section{Related Work}
\label{sec:related}
\smallsec{Robotic Manipulation in 3D}
Compared to manipulation in the 2D top-down setting~\cite{zeng:corl2020,goyal:cvpr2022,shi:corl2023}, inferring robots' movements and interactions in full 3D space is much more challenging due to the higher degrees of freedom in the action space and the complexity of 3D spatial reasoning~\cite{goyal2020packit}. To tackle manipulation in the 3D space, recent works have leveraged various perceptual representations. Camera images have been widely used for vision-based manipulation, e.g., in models such as RT-1~\cite{brohan:arxiv2022}, RT-2~\cite{zitkovich:corl2023}, and ALOHA~\cite{zhao:rss2023}. For more effective 3D spatial reasoning, depth information is commonly required, where RGB-D images are assumed as input to the manipulation policy~\cite{sundaresan:arxiv2023}. PolarNet~\cite{chen:corl2023} and M2T2~\cite{yuan:corl2023} directly use the point cloud reconstructed from RGB-D images and process it with an encoder plus a transformer to predict actions. C2F-ARM~\cite{james:cvpr2022}, PerAct \cite{shridhar:corl2022}, and FourTran~\cite{huang:iclr2024} voxelize the point clouds and use a 3D convolutional network as the backbone for action inference. Act3D~\cite{gervet:corl2023} and ChainedDiffuser~\cite{xian:corl2023} represent the scene as a multi-scale 3D feature cloud. To boost both the time efficiency and task efficacy, RVT~\cite{goyal:corl2023} proposes to use multi-view virtual images as the scene representation. Nonetheless, most of these prior models are only applied to real-world tasks that do not require high precision actions. We hereby aim to leverage these advances and push the boundary further on high precision manipulation problems. Inspired by prior work that uses a multi-stage ``coarse-to-fine'' inference strategy~\cite{james:cvpr2022,gervet:corl2023}, our model selects a task-critical part of the scene to ``zoom into'' and examine in a finer resolution through virtual images.

\smallsec{Transformers for Manipulation}
Transformer architectures have been widely adopted in robot learning for enhancing control performance~\cite{johnson:arxiv2021,chaplot:icml2021,clever:neuripsw2021,yang:iclr2022}. With the flexibility to receive heterogeneous observations as inputs, transformer-based models have emerged as powerful tools to extract features from multi-modality sensory inputs~\cite{dasari:corl2020,cachet:neuripsw2020,kim:iros2021,jangir:ral2022,liu:icra2022,zhao:rss2023,goyal:corl2023,singh2022progprompt,simeonov2023shelving}. Recent works also extend this multi-modal flexibility by integrating the transformer backbone with diffusion models to facilitate long-horizon motion planning~\cite{chi:rss2023,octo:2023}. A notable trend in prior studies is their reliance on large training datasets, often involving hundreds of demonstrations per task, to train robust transformer models. In contrast, our \method model demonstrates efficacy and proficiency in high-precision tasks with as few as 10 demos per task in real-world experiments. MimicPlay~\cite{wang2023mimicplay} is another work that attempts to learn from few demonstrations. It leverages videos of humans doing the relevant task to create a pre-trained latent representation, and is then fine-tuned with 20-40 robot demonstrations to learn a task. In contrast, RVT-2 learns directly with 10 robot demonstrations for the task. Further, MimicPlay focuses on long-horizon tasks and attempts to learn a continuous control policy, while RVT-2 focuses on learning high-precision tasks requiring millimeter-level precision, like inserting a plug in a socket and operates at the level of key-points.

\smallsec{High Precision Manipulation}
High-precision manipulation is required for tasks that have low motion error tolerance such as those in industrial settings. To learn high-precision manipulation policies, previous works have relied on various sensory modalities and data-expensive learning algorithms. As an earlier work, proprioception sensory data is used to learn a peg-in-hole policy via imitation learning~\cite{gubbi:iccar2020}. By using camera images,~\citet{schoettler:iros2020} presents a residual reinforcement learning algorithm to accomplish industrial insertion tasks from visual sensory inputs.~\citet{tang:rss2023} propose to detect the peg location from the initial camera frame and apply reinforcement learning to learn a final-inch insertion policy from proprioception data. To further improve the execution accuracy, touch feedback such as force-torque sensors~\cite{lee:icra2019,liu:arxiv2020} and vision-based tactile sensors~\cite{dong:icra2021,xu:corl2022} are exploited. However, these works leverage algorithms requiring expensive training data (e.g., either reinforcement learning or imitation learning from hundreds of demonstrations) and still only learn a single model per task. In contrast, our \method is able to learn a multi-task high-precision manipulation policy from much fewer demonstrations per task. ACT~\cite{zhao:rss2023} is another method that aims to learn precise manipulation from a few demonstrations. However,
there are significant differences between ACT and \method. Given language input,
\method can solve different variations of a task while ACT
does not take language as input and can only be trained with
one variation of a task at a time. \method makes key-point
based predictions while ACT makes continuous joint state
predictions. \method takes point cloud as input while ACT works
with multiview images.

\smallsec{Virtual Views for 3D Vision}
The use of virtual views provides a strategic lever for exploiting well-established image-based neural network architectures, such as convolutional neural networks and transformer models, for processing 3D scene information. Prior works have shown the benefit of virtual view rendering over sophisticated point-based methods in various vision tasks, from object recognition~\cite{su:iccv2015,goyal:icml2021,hamdi:iccv2021,hamdi:iclr2023}, object detection~\cite{chen:cvpr2017}, to 3D visual grounding \cite{huang:cvpr2022}. The application of virtual views in the field of robotics has been less explored. Recently, RVT \cite{goyal:corl2023} leverages multi-view representation for predicting robot actions for object manipulation. Our work builds upon RVT using a series of architectural and system-level improvements to make it more performant and efficient.

\section{Method}
\begin{figure*}[h]
    \centering
    \includegraphics[width=\textwidth,trim=82 0 0 0,clip]{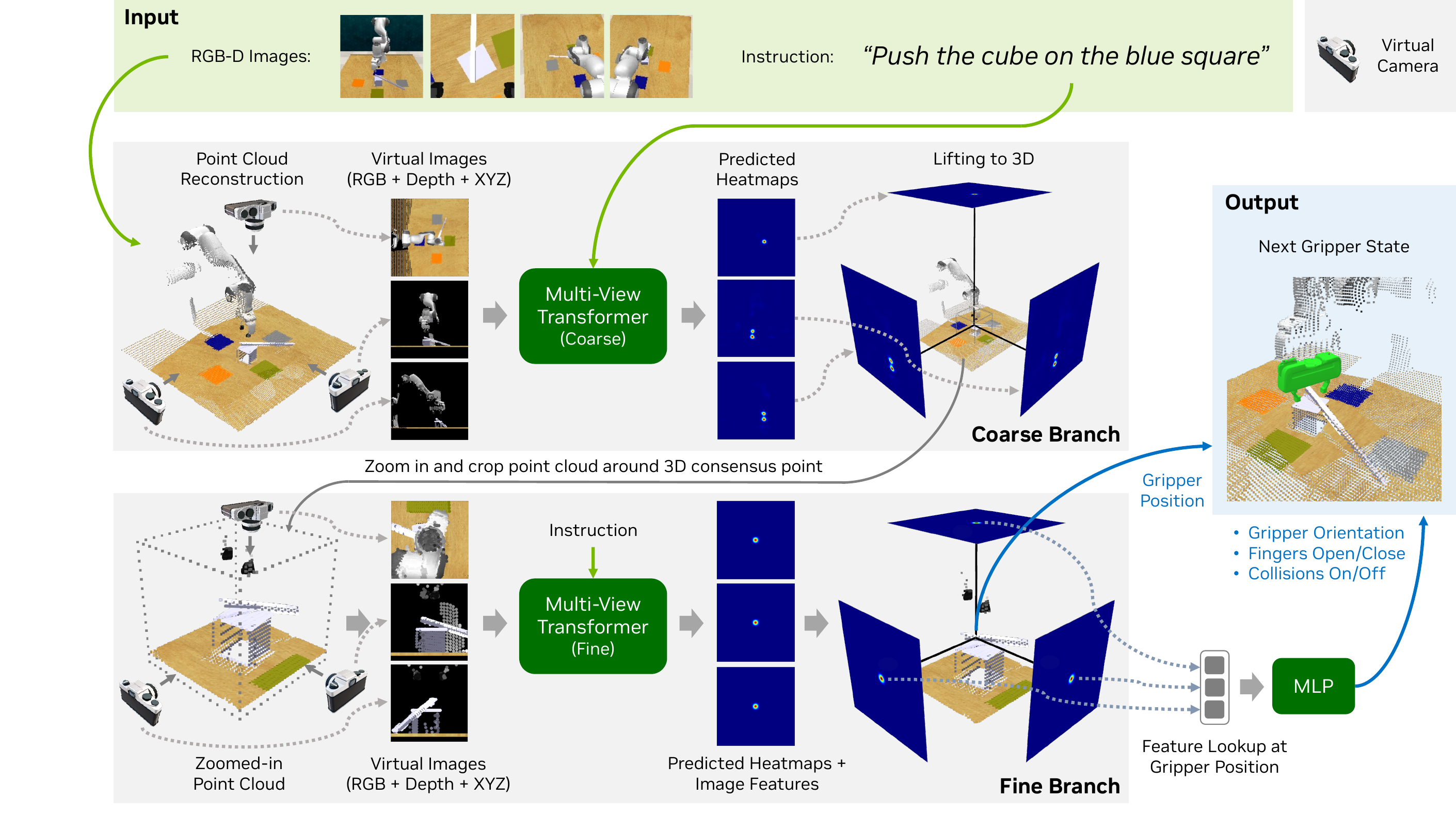} 
    \caption{\textbf{\method Architecture.} Given the current scene and a task instruction, \method predicts the next key-frame pose. It consists of two stages. The first stage uses fixed virtual views around the robot to predict the area of interest. The second stage uses zoomed-in views from the area of interest to predict the gripper pose.}
    \label{fig:overview}
    \vspace{-2mm}
\end{figure*}

Our method, \method, allows for three-dimensional and precise manipulation. A single \method model is trained to solve multiple tasks from language instructions and requires only a few demonstrations per task. \method builds upon RVT~\cite{goyal:corl2023}, a state-of-the-art model for 3D object manipulation. Similar to RVT, \method is based on the paradigm of key-frame based manipulation. But it achieves better task performance, precision, and speed through a series of improvements. We group these improvements into two categories: architectural for the ones related to changes in the neural network, and system-related for those related to software optimizations.

\subsection{Background}
\smallsec{Key-frame based manipulation}
In key-frame based manipulation, the robot trajectory is described using a sequence of key (or bottleneck) poses. For example, a trajectory for drawer opening could be described by a sequence of key poses like \textit{pre-grasp for drawer handle}, \textit{grasp pose}, and \textit{pull-pose for the drawer}. These key poses are provided in the training dataset, and the aim of a key-frame based method is to learn to predict these poses. Specifically, methods like PerAct~\cite{shridhar:corl2022} and RVT~\cite{goyal:corl2023} take as input the language goal along with the current scene point cloud and predict the next key-frame pose. The predicted pose is then passed to a motion planner, which generates a trajectory towards it\footnote{Along with the pose, these methods also output whether or not the motion planner should avoid collision.}. When the robot reaches the predicted pose, the method takes in a new scene point cloud and predicts the subsequent key-frame pose. This process iterates until the task is successful or a predefined number of steps is reached. 

To train a key-frame based behavioral cloning agent, we assume access to a dataset of samples. Each sample includes a language goal, current visual observation, and the next key-frame pose. We can extract such a dataset automatically from dense robot trajectory datasets by defining rules that specify the key-frame poses. For instance, when the state of the gripper changes between open and close, the pose is a key-frame pose. We use the same key-frame extraction scheme as PerAct and encourage readers to refer to~\citet{shridhar:corl2022} for details. 

\vspace{1mm}
\smallsec{Robotic View Transformer (RVT)}
To predict the key-frame pose, RVT first reconstructs a point cloud of the scene using the input RGB-D images. The scene is then rendered from virtual cameras along orthogonal directions. RVT renders five virtual views, including the top, front, left, back, and right view. RVT shows that using these fixed virtual views around the robot, instead of the original input camera views, results in more effective performance.

These virtual images are then passed to a multi-view transformer model that jointly reasons over all the views. The transformer model predicts a heatmap for each of the views. The heatmap score across views is then back-projected into 3D, where each 3D point receives a score that is the average of the score received by its 2D projections. The 3D point with the largest heatmap score represents the predicted gripper location. Along with the heatmaps, RVT extracts a global feature concatenated from across the views to predict the gripper rotation and state (open or close). We encourage readers to refer to \citep{goyal:corl2023} for a detailed overview of RVT.

\subsection{Architectural Changes: RVT $\to$ \method}

\vspace{1mm}
\smallsec{Multi-stage Design}
 RVT predicts the gripper pose using a fixed set of views around the robot. These fixed views might not be sufficient in tasks when the object of interest is very small, and the gripper pose needs to be very precise, like \textit{inserting a peg in a hole}. Hence, \method adopts a multi-stage design (see Fig.~\ref{fig:overview}), where, in the first or coarse stage, it predicts the area of interest using a fixed set of views. \method then zooms in the area of interest and re-renders images around it. We use a zoom-in factor of 4, meaning that zoomed-in cameras cover a region of size $1/ 4^{th}$ the coarse cameras. It uses these close-up images to make precise gripper pose predictions. Such adaptive rendering is possible due to the flexibility provided by the virtual rendering proposed in RVT.

\vspace{1mm}
\smallsec{Convex Upsampling}
RVT is based on ViT (Vision Transformer)~\cite{dosovitskiy:iclr2021}, which divides an image into $t_1 \times t_2$ patches. Each image patch is processed as a single token of dimension $d$. To predict a heatmap from the image tokens, RVT first arranges the image tokens at their corresponding patch location. This results in a 3D feature of shape $t_1 \times t_2 \times d$. RVT then uses transposed convolutions to upsample these features to the image resolution of $h \times w$, creating a feature of shape $h \times w \times d$. Finally, these features of shape $h \times w \times d$ are used to predict the heatmap. This sequence of operations is effective, however, it is memory intensive due to the large intermediate feature of shape $h \times w \times d$. 

To address this, \method removes the feature upsampling and directly predicts heatmap of shape $h \times w$ from features at the token resolution. Specifically, it uses convex upsampling layer proposed by~\cite{teed:eccv2020}. The convex upsampling layer uses a learned convex combination of features in the coarse grid to make predictions in the higher resolution. \cite{teed:eccv2020} shows how it leads to sharper predictions at higher resolution. We empirically find that convex upsampling saves memory without sacrificing performance (see Tab.~\ref{table:rlbench_ablation}). The convex upsampling layer does not require any special implementation and can be represented using the native \texttt{fold} function in PyTorch.

\vspace{1mm}
\smallsec{Parameter Rationalization}
We find that network parameters in RVT, like the virtual image size (220) and patch size (11) may not be optimal for GPU as they are not divisible by powers of 2 like 16\footnote{Exact power of 2 depends on the data-type and NVIDIA GPU architecture. More details can be found at \url{https://docs.nvidia.com/deeplearning/performance/dl-performance-fully-connected/index.html}.}. \method rationalizes these parameters to make the neural network more GPU-friendly, improving its speed. \method adopts parameters similar to ViT~\cite{dosovitskiy:iclr2021}, i.e. image size of 224 and patch size of 14. Apart from being more GPU-friendly, these parameters reduce the total number of tokens inside the multi-view transformer which is equal to $(image\_size / patch\_size)^2$, boosting the speed further. These choices make \method faster during training and testing without affecting performance (see Tab.~\ref{table:rlbench_ablation}).
 
\vspace{1mm}
\smallsec{Location Conditioned Rotation}
RVT and PerAct use global visual features, like max-pooling over the entire image, to make predictions for end-effector rotation. This can be problematic when there are multiple valid end-effector locations, and the end-effector rotation depends on the location. For example, consider the task of stacking blocks where the scene has two similar blocks but in different orientations. Here, picking either of the two blocks is a valid step. However, since the blocks have different orientations, the end-effector rotation would depend on the chosen end-effector location. Since RVT only uses global visual features to predict rotation, it cannot handle such cases. To address this, \method uses local features pooled from the feature map at the end-effector location for rotation prediction. This allows \method to make location-dependent rotation prediction.

\vspace{1mm}
\smallsec{Fewer Virtual Views}
RVT renders the scene point cloud with five virtual cameras placed in orthogonal locations i.e. back, front, top, left, and right. This choice was based on their observation that fewer camera views reduced performance. However, in our multi-stage \method model, we find that using only three views, i.e., front, top, and right, suffices and does not sacrifice performance. This is likely because \method uses zoomed-in views for the final prediction. Fewer virtual views reduce the number of images to be rendered by the renderer and the number of tokens to be processed by the multi-view transformer. Thus, this improves training and inference speed.

\subsection{System-Related Changes: RVT $\to$ \method}
\vspace{1mm}
\smallsec{Point-Renderer}
RVT uses PyTorch3D~\cite{ravi:arxiv2020} to render virtual RGB-D images. PyTorch3D is an appealing choice because of its easy-to-use interface. However, it is a fully-featured differentiable renderer that incurs significant time and memory overhead for point-cloud rendering. To avoid this, we implement a custom projection-based point-cloud renderer in CUDA.
Our renderer performs 3 steps to render a point cloud with $N$ points to an RGB image and depth image of size $(h,w)$:

\paragraph{Projection}
For each 3D point of index $n \in \{0, 1 \dots N\}$ and RGB value $f_n$, it computes the depth $d_{n}$ and image pixel coordinate $(x_{n}, y_{n})$ using camera intrinsics and extrinsics. From the 2D pixel coordinate $(x_{n}, y_{n})$, it computes the linear pixel index $i_{n} = x_{n} \cdot w + y_{n}$. The projection operation is easily accelerated using GPU matrix multiplications.

\paragraph{Z-ordering}
For each pixel of linear-index $j$ in the image, it finds the point index with smallest depth $d_{n}$ among the set of points that project to the pixel $\{n \;|\; i_{n} = j\}$. It assigns that point's RGB value $f_{n}$ to pixel $j$ of the RGB image and depth $d_{n}$ to pixel $j$ of the depth image.

To accelerate Z-ordering, we pack each point's depth and index into a single 64-bit integer, such that the most significant 32 bits encode depth, while the least significant bits encode the point index. Then, Z-ordering can be implemented with two CUDA kernels. First, a parallel loop over point cloud points, tries to store each packed depth-index into a depth-index image at the pixel $j$ using the \textit{atomicMin} operation. Only the depth-index stored by the minimum-depth point at each pixel survives. The second kernel, in a loop over pixels, creates depth and feature images by unpacking the depth-index, and looking up the point feature. This trick was proposed by \citet{schutz:cgf2021} for rendering color point-clouds by packing the 32-bit color. We extend this to images with arbitrary number of channels, by packing the point index instead.

\paragraph{Screen-space splatting}
The first two steps are sufficient to produce rendered images. However, the points are treated as infinitesimal light sources, which creates noise in areas where the screen-space point cloud resolution is not higher than the image resolution. A common way to counter this is 3D splatting, whereby each point is modelled by some geometry of a finite size. We represent each point as a disc of radius $r$ facing the camera. This splatting can be computed in screen space after projection and z-ordering, thereby reducing the computation required in the projection and z-ordering.
For each pixel $j$ in the image, search in a neighbourhood for another pixel $k$ of lowest depth. If the pixel $k$ has depth $d_{k} < d_{j}$, and is closer than $r \cdot focal\_length / d_{k}$, replace the feature and depth of pixel $j$ with that of pixel $k$. 

\vspace{1mm}
\smallsec{Improved training pipeline}
We optimize RVT's training pipeline by adopting the latest developments in training transformers. We analyze various techniques and adopt the ones that improve speed without affecting performance. Specifically, we use mixed precision training, 8-bit LAMB optimizer~\cite{dettmers:iclr2022}, and fast GPU implementation of the attention layer based on xFormers~\cite{lefaudeux:2022}.

\section{Experiments}
\label{sec:exp}
We evaluate \method by conducting comprehensive experiments in both the simulation and the real-world.

\subsection{Simulation}
\smallsec{Dataset and Setup}
We conduct the experiments on a standard multi-task manipulation benchmark developed in RLBench \cite{james:ral2020} and adopted by previous works \cite{shridhar:corl2022,gervet:corl2023,goyal:corl2023}. The benchmark contains $18$ tasks, including non-prehensile tasks like \textit{push buttons}, common pick-and-place tasks like \textit{place wine}, and peg-in-hole tasks that require high precision like \textit{insert peg}. Each task is specified by a language description and consists of $2$ to $60$ variations such as handling objects in different colors or locations. A Franka Panda robot with a parallel jaw gripper is commanded to complete the tasks. The task and the robot are simulated via CoppelaSim~\cite{rohmer:iros2013}. The input RGB-D images are of resolution $128\times 128$ and are captured by four noiseless cameras mounted at the front, left shoulder, right shoulder, and wrist of the robot. We train and test \method with the same dataset as PerAct and RVT, with $100$ demonstrations per task for training and $25$ unseen demonstrations for testing\footnote{for the close jar task, we use the success criteria fixed by Tsung-Wei Ke here: \url{https://github.com/buttomnutstoast/RLBench/commit/587a6a0e6dc8cd36612a208724eb275fe8cb4470}. The fix is used in Act3D. This fix did not affect the performance of the released RVT~\cite{goyal:corl2023}.}

\vspace{1mm}
\smallsec{Training and Evaluation Details}
We train \method with similar computing resources as RVT and PerAct. Specifically, we use a node with 8 NVIDIA V100 16 GB GPUs. Like RVT and PerAct, we use translation augmentation of 12.5 cm along the $x$, $y$, and $z$ axis, as well as rotation augmentation of $45^{\circ}$ along the $z$ axis. We train  \method for $\sim$80K steps with a cosine learning rate decay schedule and an initial warmup of 2000 steps. The batch size is 192 (24 $\times$ 8) and the learning rate is $2.4 \times 10^3$. We use the final model for evaluation. Since RLBench uses a sampling-based motion planner, we evaluate each model four times on each task and report the mean and variance. We measure the inference speed of \method, RVT, and PerAct on an NVIDIA RTX 3090 GPU.

\begin{table*}[t]
\centering
\Huge
\resizebox{\textwidth}{!}{
\begin{tabular}{lcccccccccccccccccc} 
\toprule
\rowcolor[HTML]{CBCEFB}
                                                & Avg.                 & Avg.                & Train time             &  Inf. Speed          & Close               & Drag                  & Insert              & Meat off            & Open                & Place                & Place \\
\rowcolor[HTML]{CBCEFB}
Models                                          & Success $\uparrow$   & Rank $\downarrow$   & (in days) $\downarrow$ &  (in fps) $\uparrow$ & Jar                 & Stick                 & Peg                 & Grill               & Drawer              & Cups                 & Wine  \\
\midrule
\bczcnn~\cite{jang:corl2021,shridhar:corl2022}  & ~~1.3                & 7.4                 & -                      & -                    & 0                   & 0                     & 0                   & 0                   & 4                   & 0                    & 0 \\
\rowcolor[HTML]{EFEFEF}
\bczvit~\cite{jang:corl2021,shridhar:corl2022}  & ~~1.3                & 7.7                 & -                      & -                    & 0                   & 0                     & 0                   & 0                   & 0                   & 0                    & 0 \\
\unet~\cite{james:cvpr2022,shridhar:corl2022}   & 20.1                 & 5.8                 & -                      & -                    & 24                  & 24                    & 4                   & 20                  & 20                  & 0                    & 8 \\
\rowcolor[HTML]{EFEFEF}
HiveFormer~\cite{guhur:corl2022}                & 45.3                 & 5.2                 & -                      & -                    & 52.0                & 76.0                  & 0.0                 & \tb{100.0}          & 52.0                & 0.0                  & 80 \\
PolarNet~\cite{chen:corl2023}                   & 46.4                 & 4.8                 & -                      & -                    & 36.0                & 92.0                  & 4.0                 & \tb{100.0}          & 84.0                & 0.0                  & 40 \\
\rowcolor[HTML]{EFEFEF}
\peract~\cite{shridhar:corl2022}                & 49.4                 & 4.4                 & 16.0                   & 4.9                  & 55.2 \rpmh 4.7      & 89.6 \rpmh 4.1        & 5.6 \rpmh 4.1       & 70.4 \rpmh 2.0      & 88.0 \rpmh 5.7      & 2.4 \rpmh 3.2        & 44.8 \rpmh 7.8 \\
Act3D~\cite{gervet:corl2023}                    & 65.0                 & 2.8                 & 5.0                    &                      & 92.0           & 92.0                  & 27.0                & 94.0                & \tb{93.0}           & 3.0                  & 80 \\
\rowcolor[HTML]{EFEFEF}
RVT~\cite{goyal:corl2023}                       & 62.9                 & 2.8                 & 1.0                    & 11.6                 & 52.0      \rpmh 2.5 & \tb{99.2} \rpmh 1.6   & 11.2 \rpmh 3.0      & 88.0 \rpmh 2.5      & 71.2 \rpmh 6.9      & 4.0 \rpmh 2.5         & 91.0 \rpmh 5.2 \\
\method (ours)                                  & \tb{81.4}            & \tb{1.5}            & \tb{0.83}              & \tb{20.6}            & \tb{100.0} \rpmh 0.0 & 99.0 \rpmh 1.7        & \tb{40.0} \rpmh 0.0 & 99.0 \rpmh 1.7      & 74.0 \rpmh 11.8     & \tb{38.0} \rpmh 4.5   & \tb{95.0} \rpmh 3.3 \\
\midrule
\rowcolor[HTML]{CBCEFB}
                                                & Push                 & Put in              & Put in                  & Put in               & Screw               & Slide                 & Sort                & Stack               & Stack               & Sweep to             & Turn \\
\rowcolor[HTML]{CBCEFB}
Models                                          & Buttons              & Cupboard            & Drawer                  & Safe                 & Bulb                & Block                 & Shape               & Blocks              & Cups                & Dustpan              & Tap \\
\midrule
\bczcnn~\cite{jang:corl2021,shridhar:corl2022}  &  0                   & 0                   & 8                       & 4                    & 0                   & 0                     & 0                   & 0                   & 0                   & 0                    & 8  \\
\rowcolor[HTML]{EFEFEF}
\bczvit~\cite{jang:corl2021,shridhar:corl2022}  &  0                   & 0                   & 0                       & 0                    & 0                   & 0                     & 0                   & 0                   & 0                   & 0                    & 16 \\
\unet~\cite{james:cvpr2022,shridhar:corl2022}   & 72                   & 0                   & 4                       & 12                   & 8                   & 16                    & 8                   & 0                   & 0                   & 0                    & 68 \\
\rowcolor[HTML]{EFEFEF}
HiveFormer~\cite{guhur:corl2022}                & 84                   & 32.0                & 68.0                    & 76.0                 & 8.0                 & 64.0                  & 8.0                 & 8.0                 & 0.0                 & 28.0                 & 80  \\
PolarNet~\cite{chen:corl2023}                   & 96                   & 12.0                & 32.0                    & 84.0                 & 44.0                & 56.0                  & 12.0                & 4.0                 & 8.0                 & 52.0                 & 80  \\
\rowcolor[HTML]{EFEFEF}
\peract~\cite{shridhar:corl2022}                & 92.8 \rpmh 3.0       & 28.0 \rpmh 4.4      & 51.2 \rpmh 4.7          & 84.0 \rpmh 3.6       & 17.6 \rpmh 2.0      & 74.0 \rpmh 13.0       & 16.8 \rpmh 4.7      & 26.4 \rpmh 3.2      & 2.4 \rpmh 2.0       & 52.0 \rpmh 0.0       & 88.0 \rpmh 4.4 \\
Act3D~\cite{gervet:corl2023}                    & 99                   & 51.0                & 90.0                    & 95.0                 & 47.0                & \tb{93.0}             & 8.0                 & 12.0                & 9.0                 & 92.0                 & 94                  \\
\rowcolor[HTML]{EFEFEF}
RVT~\cite{goyal:corl2023}                       & \tb{100.0} \rpmh 0.0 & 49.6 \rpmh 3.2      & 88.0 \rpmh 5.7          & 91.2 \rpmh 3.0       & 48.0 \rpmh 5.7      & 81.6 \rpmh 5.4        & \tb{36.0} \rpmh 2.5 & 28.8 \rpmh 3.9      & 26.4 \rpmh 8.2      & 72.0 \rpmh 0.0      & 93.6 \rpmh 4.1 \\
\method (ours)                                  & \tb{100.0} \rpmh 0.0 & \tb{66.0} \rpmh 4.5 & \tb{96.0} \rpmh 0.0     & \tb{96.0} \rpmh 2.8  & \tb{88.0} \rpmh 4.9 & 92.0 \rpmh 2.8        & 35.0 \rpmh 7.1      & \tb{80.0} \rpmh 2.8 & \tb{69.0} \rpmh 5.9 & \tb{100.0} \rpmh 0.0 & \tb{99.0} \rpmh 1.7 \\
\bottomrule
\end{tabular}
}
\caption{\textbf{Multi-Task Performance on RLBench.} We report the success rate for 18 RLBench~\cite{james:ral2020} tasks and the average success rate across all the tasks. The success condition is as defined in RLBench. \method outperforms all methods while having higher training and inference speed. Performance of HiveFormer and PolarNet are reported by \citep{chen:corl2023}; RVT and PerAct are reported by \cite{goyal:corl2023}; and Act3D is reported by \citep{gervet:corl2023}. All are trained with 100 demonstrations and a single model is evaluated on all the tasks. All methods use input images of resolution $128 \times 128$, except Act3D, which uses $256 \times 256$.
}
\vspace{-4mm}
\label{table:rlbench}
\end{table*}

\begin{figure}[t]
    \centering
    \includegraphics[width=\columnwidth]{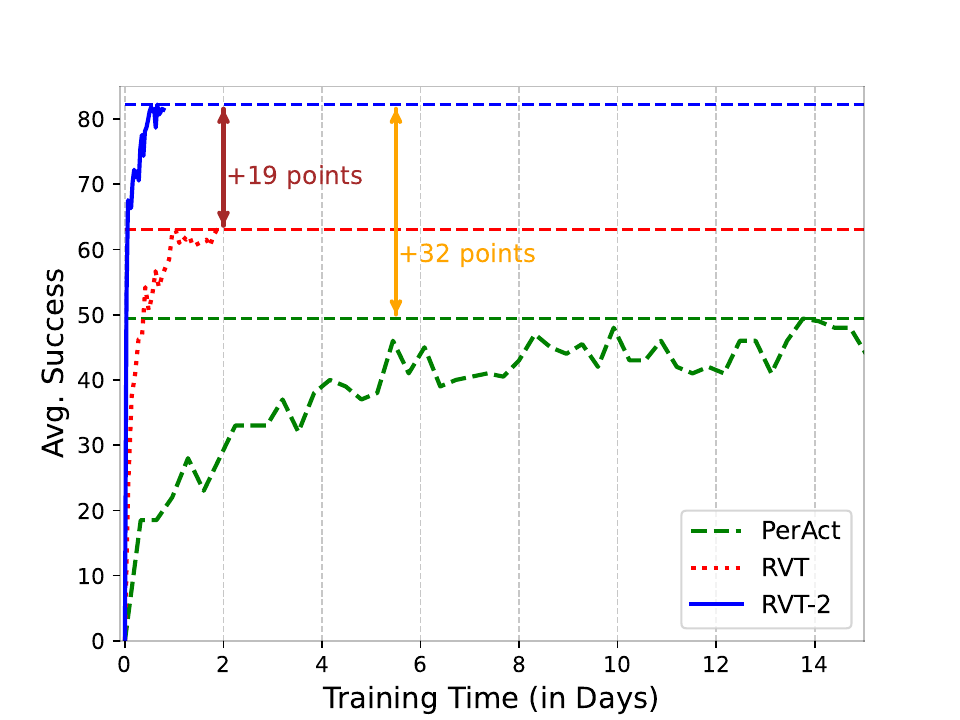} 
    \caption{\textbf{Training time vs Success rate on RLBench.} All models are trained on 8 NVIDIA V100 GPUs. \method trains significantly faster and achieves higher performance than prior state-of-the-art RVT and PerAct.}
    \vspace{-2mm}
    \label{fig:plot}
\end{figure}

\vspace{1mm}
\smallsec{Baselines}
We compare \method with various baselines. These include simple image-to-action behavioral cloning baselines, \textit{Image-BC (CNN)}~\cite{jang:corl2021,shridhar:corl2022} and \textit{Image-BC (ViT)}~\cite{jang:corl2021,shridhar:corl2022}, that use a CNN and ViT backbone respectively. We also compare with models that have been specifically designed for 3D object manipulation including \textit{C2F-ARM-BC}~\cite{james:ral2020}, \textit{PerAct}~\cite{shridhar:corl2022} and \textit{HiveFormer}~\cite{guhur:corl2022}; as well as more recently proposed methods like \textit{RVT}~\cite{goyal:corl2023}, \textit{PolarNet}~\cite{chen:corl2023} and \textit{Act3D}~\cite{gervet:corl2023}. All baselines and \method are trained and tested with input images of $128 \times 128$, while Act3D uses images of size $256 \times 256$. 

\vspace{1mm}
\smallsec{Training time vs. Performance} In Fig.~\ref{fig:plot}, we compare the training time and success rate on RLBench for \method, RVT, and PerAct. We find that \method significantly outperforms both while requiring much less compute to train. Because of the efficiency gains, with the same compute, \method trains 6X faster\footnote{
RVT can fit a batch size of 24 and train for 100K steps, equivalent to training over 2.4M samples in 24 hours. \method fits a batch size of 192 and trains for 83.3k steps, equivalent to 16M samples in 20 hours.} than RVT, while improving performance by 19\% in absolute or 29\% in relative terms. While comparing with PerAct, \method improves the relative performance by 65\%. We find that within 2 hours of training, \method outperforms RVT trained for 24 hours and PerAct trained for 16 days. These efficiency gains could allow for further scaling up \method in the future. In inference speed, \method exhibits around 2X improvement compared to RVT. With an inference speed of 20 fps, \method opens up new possibilities for real-time reactive control.

\vspace{1mm}
\smallsec{Mulit-Task Performance}
Table \ref{table:rlbench} summarizes the comparison of RVT-2 with prior methods on the RLBench tasks. Among all the methods, \method achieves the highest average success rate of $81.4\%$. \method outperforms the prior best-performing model Act3D by $16.4\%$ absolute or $25\%$ relative improvement while requiring 6X less compute to train (5 days vs. within 20 hours). From Fig.~\ref{fig:plot}, we see that \method achieves higher performance than Act3D with just 4 hours of training. Overall, \method achieves the best results in $13$ out of $18$ tasks and an average rank of $1.5$. 

Out of the 18 tasks, the task where \method does not achieve close to the best results is \textit{open drawer}. Upon further investigation, we find that on \textit{open drawer}, \method achieves higher success rates like 86\% on earlier checkpoints than the final one. The lower performance on the final checkpoint could be an artifact of over-fitting or multi-task training where performance on some tasks degrades while improving on others.

\vspace{1mm}
\smallsec{High-Precision Tasks} We find that \method outperforms other methods on high-precision tasks like \textit{insert peg}, \textit{stack cups} and \textit{screw bulb}. In \textit{insert peg}, the robot must pick up a square peg on the tabletop and insert it onto a specific cuboid stick. This task can effectively examine the precision of the learned model since the clearance between the stick and the peg is very tight, and the robot has to align the square peg perfectly with the cuboid stick; otherwise, any tiny error will result in a failure insertion. In \textit{stack cups}, a minor error in the pick and place locations of the cup results in failure as seen in the low success rate of prior methods. Similarly, in \textit{screw bulb}, the bulb's base must be well aligned with the socket for successful screwing. Our experiments show that \method achieves a significantly higher success rate on these tasks, achieving 88\% versus 48\% for the previous best on \textit{scew bulb}; 69\% versus 26.4\% on \textit{stack cups} and 40\% versus 27\% on \textit{insert peg}.

\begin{figure}[t]
 \centering
 \includegraphics[width=0.89\linewidth]{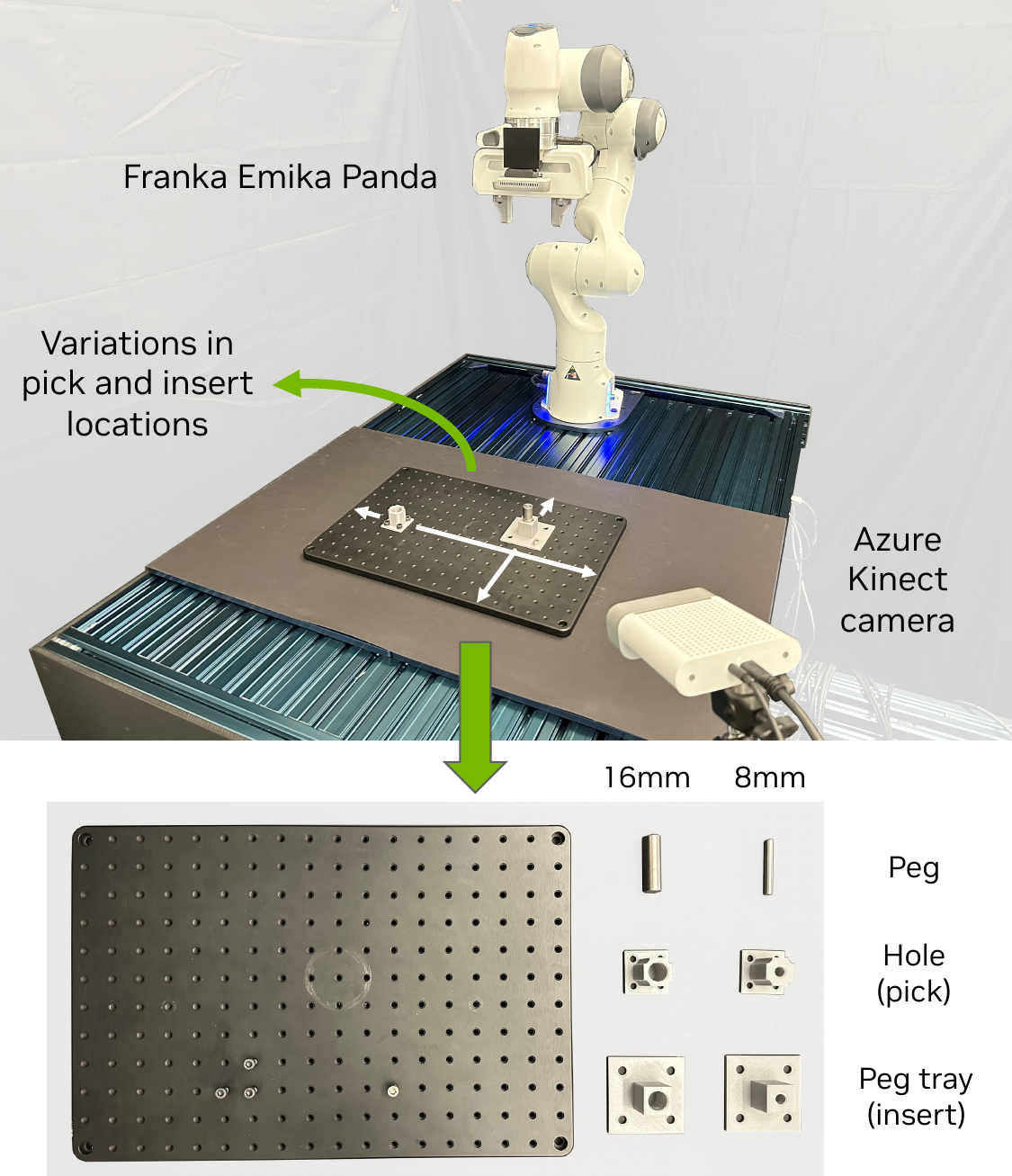}
 \caption{Our real-world setup (top) and the peg picking and insertion task from~\cite{tang:rss2023}.}
 \label{fig:setup_1}
\end{figure}

\begin{figure}[!]
 \centering
 \includegraphics[width=\linewidth]{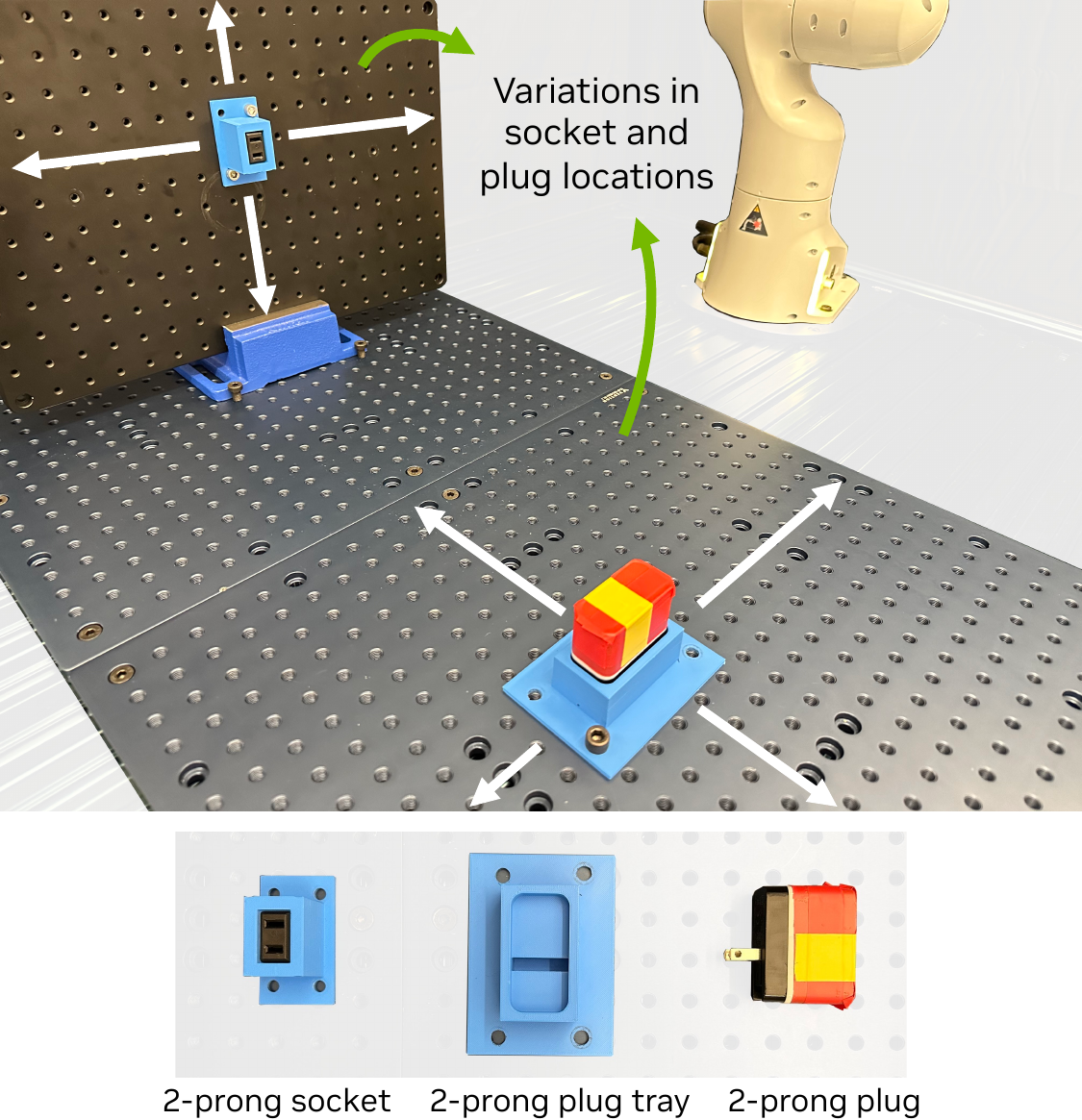}
 \caption{The 2-prong plug picking and insertion task from~\cite{tang:rss2023}; and the considered variations in object locations.}
 \vspace{-2mm}
 \label{fig:setup_2}
\end{figure}

\begin{table}[tb]
 \centering
 \setlength\tabcolsep{2pt}
 \resizebox{\columnwidth}{!}{
 \begin{tabular}{lccccccc}
\toprule
\rowcolor[HTML]{CBCEFB}
                                         & \# of      & \# of   & \# of & \multicolumn{2}{c}{\textit{Models}}           \\
\rowcolor[HTML]{CBCEFB}
Task                                     & vari.      & train   & test   & RVT                       & \method (ours) \\
\midrule
Stack  blocks                            & ~~3        & 15      & 10     & \tb{80\%}                 & \tb{80\%}       \\
\rowcolor[HTML]{EFEFEF}
Press sanitizer                          & ~~1        & 7       & 10     & \tb{90\%}                 & 80\%            \\
Put marker in mug/bowl                   & ~~4        & 12      & 10     & 20\%                      & \tb{50\%}       \\
\rowcolor[HTML]{EFEFEF}
Put object in drawer                     & ~~3        & 12      & 10     & 30\%                      & \tb{50\%}       \\
Put object  in shelf                     & ~~2        & 8       & 10     & \tb{100\%}                & \tb{100\%}      \\
\midrule
All tasks in RVT~\cite{goyal:corl2023}   & 13         & 54      & 50     & 64\%                      & \tb{72\%}       \\
\midrule
\rowcolor[HTML]{EFEFEF}
Pick and insert 16mm peg                 & ~~1        & 10      & 10     & ~~50\%                    & \tb{60\%}        \\
Pick and insert 8mm peg                  & ~~1        & 10      & 10     & ~~40\%                    & \tb{50\%}        \\
\rowcolor[HTML]{EFEFEF}
Pick and insert plug                     & ~~1        & 10      & 10     & ~~10\%                    & \tb{50\%}       \\
\midrule
All high precision tasks                 & 3          & 30      & 30     & ~~33.3\%                  & \tb{53.3\%}        \\
\midrule
All tasks                                & 16         & 84      & 80     & ~~52.5\%                  & \tb{65\%}        \\
\bottomrule
 \end{tabular}
 }
 \caption{\textbf{Results in the real world.} Both \method and RVT use a single model for all 8 tasks with 16 variations. \method outperforms RVT on the tasks from~\cite{goyal:corl2023} and the new high-precision tasks.}
 \vspace{-3mm}
 \label{table:real}
\end{table}

\subsection{Real World}
\smallsec{Dataset and Setup}
We compare \method with RVT on a real-world manipulation setup similar to that used in RVT (Fig.~\ref{fig:setup_1} top). The setup consists of a statically mounted Franka Emika Panda arm and a static third-person view Azure Kinect RGB-D camera. Instead of following the camera position in RVT, we move the camera closer to the robot's workspace to ensure the point cloud's quality for high-precision manipulation. We use this camera position for all tasks. Besides the same five tasks used in RVT (\textit{stack blocks}, \textit{press sanitizer}, \textit{put marker in mug/bowl}, \textit{put object in drawer}, \textit{put object in shelf}), we additionally evaluate on three high-precision tasks from IndustRealKit~\cite{tang:rss2023}: \textit{pick and insert 16mm peg}, \textit{pick and insert 8mm peg}, \textit{pick and insert plug}. The two peg tasks consist of picking up the peg from a hole and inserting it into another hole (Fig.~\ref{fig:setup_1}). Both the pick and place locations are randomized over the work surface during evaluation. The plug task consists of picking up a 2-prong plug from a tray and inserting it into a vertically mounted socket (Fig.~\ref{fig:setup_2}). This task further goes beyond 2D pick-and-place and requires precise manipulation in the 3D space. Similarly, the location of the plug tray and socket are randomized over their work surface during evaluation. For tasks in RVT~\cite{goyal:corl2023}, we collect the same number of demonstrations ($\sim10$) as reported by them. For new high-precision tasks, we collect 10 demonstrations per task. The dataset statistics are provided in Tab.~\ref{table:real}.

\begin{table*}[t]
 \centering
 \setlength\tabcolsep{3pt}
 \begin{tabular}{ccccccccccccccccccc}
\rowcolor[HTML]{CBCEFB}
Row & Multi- & Parameter        & Loc. Cond.        & Point   & Convex   & Mixed & 8-bit Opt.     & \# of & Training  Time   & Training Time   & Avg.  & Avg. Succ.\\
\rowcolor[HTML]{CBCEFB}
ID & Stage & Rational.        & Rot               & Render  & Upsamp.    &  Prec. & + Fast Attn.   & Views  & (in hours)       & \% of base      & Succ.  & diff. wrt.  base\\
\toprule
1  & \yes   & \yes             & \yes              & \yes    & \yes      & \yes   & \yes           & 3      & 19.5           &      100\%        & 81.4  & 0\\
\rowcolor[HTML]{EFEFEF}
2  & \no    & \yes             & \yes              & \yes    & \yes      &  \yes  & \yes           & 3      & 13.0           &      67\%        & 63.9  & - 17.5\\
3  & \yes   & \no              & \yes              & \yes    & \yes      &  \yes  & \yes           & 3      & 26.7           &      137\%        & 77.2  & - 4.2\\
\rowcolor[HTML]{EFEFEF}
4  & \yes   & \yes             & \no               & \yes    & \yes      &  \yes  & \yes           & 3      & 19.3           &       99\%        & 78.9  & -2.5\\
5  & \yes   & \yes             & \yes              & \no     & \yes      &  \yes  & \yes           & 3      & 71.1           &       366\%       & 79.3  & -2.1\\
\rowcolor[HTML]{EFEFEF}
6  & \yes   & \yes             & \yes              & \yes    & \no       &  \no   & \yes           & 3      & 79.2           &      406\%        & 82.0  & +0.6\\
7  & \yes   & \yes             & \yes              & \yes    & \yes      &  \no   & \yes           & 3      & 58.5           &      300\%        & 81.2 & -0.2 \\
\rowcolor[HTML]{EFEFEF}
8  & \yes   & \yes             & \yes              & \yes    & \yes      &  \no   & \no            & 3      & 62.4           &      320\%        & 81.3 & -0.1 \\
9  & \yes   & \yes             & \yes              & \yes    & \yes      &  \yes  & \yes           & 5      & 40.3           &      207\%        & 79.7  & -1.7 \\
\bottomrule
 \end{tabular}
 ~~~~~
 \caption{\textbf{Ablations on RLBench.} We quantify the impact of all the architectural and system-level improvements in \method. All these contribute to increasing the training speed, inference speed and performance of \method.}
 \vspace{-2mm}
 \label{table:rlbench_ablation}
\end{table*}

\vspace{1mm}
\smallsec{Training and Evaluation Details}
We train both RVT and \method on the same dataset for fairness. We train a single RVT and \method model for all eight tasks. Both models are trained for 10 epochs using a cosine learning rate schedule and the same data augmentation as in our simulation experiments. For \method we use the same batch size and learning rate as in the simulation experiments. For RVT, we cannot fit a larger batch size as \method in memory, so we use the official training parameters of batch size 24 and learning rate $2.4 \times 10^3$~\cite{goyal:corl2023}. We use the final model for evaluation.

\vspace{1mm}
\smallsec{Experiment Results}
Table \ref{table:real} shows the results in the real world. We find that \method can perform multiple tasks with only a handful of demonstrations ($\sim10$) per task. Of the five tasks from RVT~\cite{goyal:corl2023}, \method outperforms RVT by 8 absolute points and 12.5 in relative terms. On all three new tasks that require high precision, \method constantly outperforms RVT and achieves $53.3\%$ average success rate versus $33.3\%$ for RVT. Although \method achieves encouraging results on the high-precision tasks with just a single camera, a common reason for failure was small errors during insertion. We believe augmenting \method with a reactive policy to make fine adjustments in the final stages of insertion could be an exciting future direction. We encourage readers to view video results provided on the project website for success and failure examples.
\vspace{1mm}

\smallsec{\noindent Failure Modes}
We conduct a study of all failure cases across all tasks in the real world and report failure modes. We classify each failed episode into a “mode” that signifies the reason for failure. For each task, we report the percentage of failure because of a particular mode out of all the failures.\\
\vspace{-1em}
\begin{itemize}\setlength\itemsep{-0.75em}
\item \textit{Stack blocks:} Placement on the incorrect block (100\%) \\
\item \textit{Press sanitizer:} Missed sanitizer top (100\%) \\
\item\textit{Put marker in mug/bowl:} Picking the wrong marker (80\%); Not going to the goal (20\%) \\
\item\textit{Put object in drawer:} Picking wrong marker (40\%); Not going to the goal (20\%); Failure while grasping (40\%) \\
\item\textit{Put object in shelf:} No failure \\ 
\item\textit{Pick and insert 16mm peg:} Small error while placing peg (75\%); Not going to the goal (25\%) \\ 
\item\textit{Pick and insert 8mm peg:} Small error while placing peg (40\%); Not going to the goal (40\%); Failure to grasp peg (20\%) \\ 
\item\textit{Pick and insert plug:} Small error while plugging into the socket (100\%) 
\end{itemize}
Overall, minor inaccuracies in the position prediction is the major reason for failure in the ``Press sanitizer", ``Pick and insert 16mm peg", ``Pick and insert 8mm peg," and ``Pick and insert plug." For the ``Put marker in mug/bowl" and ``Put object in drawer," picking up the incorrect colored marker is a major failure mode. This could be because the markers are thin structures with few points in the point cloud informing about the color of the marker. Further, failure to grasp the object contributed to 40\% of the failed episodes in the ``Put object in drawer" and 20\% of the failed episodes in the ``Pick and insert 8mm peg" task.

\subsection{Ablations}
\label{sec:ablation}
We conduct an extensive ablation study in simulation to analyze the effect of each component of \method. Results are shown in Table \ref{table:rlbench_ablation}. 

\paragraph{Multi-Stage Design}
Comparing row 1 and 2, we can see that including multi-stage design introduces a slowdown in the training time due to the extra stage of rendering and inference. However, it brings a $17.5\%$ success rate improvement since the zoom-in view provides more task-relevant details about the region of interest.

\paragraph{Parameter Rationalisation} 
Comparing row 1 and 3, we see employing GPU-friendly network parameters accelerates the training process without compromising the performance of the network.

\paragraph{Location Conditioned Rotation} From row 1 and 4, we see that predicting the rotation using the local features improves performance by 4.2\%. 

\paragraph{Fewer Virtual Views}
We vary the number of views in row 1 and 9. It reveals that reducing the number of camera views from 5 to 3 not only maintains task performance but also leads to a twofold increase in training speed. Unlike RVT, the multi-stage network in \method performs well even with fewer virtual views per stage.

\paragraph{Convex Upsampling}
On comparing row 6 and 7, we find that removing convex upsampling increases the training time by 20.7 hours. We observe that removing convex upsampling but keeping the mixed precision leads to undefined gradients during training. Hence, to ablate convex upsampling, we compare row 6 and 7, both without mixed precision.

\paragraph{Point-Renderer}
By replacing PyTorch3D with our customized Point-Rederer, the forward pass is significantly sped up, resulting in $3.6$X faster training as seen in row 5.

\paragraph{Improved Training Pipeline}
Comparing rows 1 and 7, we see that removing automatic mixed precision training leads to a 300\% increase in training time. Further removing the 8-bit LAMB optimization and fast attention (row 8) further increases the training time by 20\%. Although the improvements due to 8-bit LAMB optimizer and fast attention are not large, they are helpful for the most optimized pipeline and could potentially be beneficial for scaling up \method.

\subsection{Generalization Case Study} 
Similar to prior works, we test generalization to unseen environment configurations. The object configurations are different in training and testing, and object positions vary in a workspace of dimension 2 feet X 2 feet.
We also test \method on other generalization scenarios, where we vary lighting conditions, background (table) appearance, and objects' appearance. Specifically, we test the block stacking task and find that even when trained with few demonstrations, \method demonstrates generalization to unseen lighting conditions, background, and modifications to the objects' appearance. 

The video of this test can be found on the project website (\purl). We also investigate the generalization to language input. For block stacking, the language input in the training dataset is of the format ``put x block on y block" where x and y are different colors. We find our model to be robust to language inputs like ``move x block such that it is on y block," ``move x block onto y block," ``stack x block on y block," and ``move blocks such that x is under y."

\subsection{Failure Recovery Case Study} 
\method uses the observation from the current time step to predict the pose in the next time step. Hence, the network is closed loop at the frequency of key-points. To demonstrate this, we do a study for the stack block task where we move the target block mid-way of the execution and find that the
policy adjusts accordingly. Please see the video of this test on the project website.

We further examined various episodes and found failure recovery behavior across several tasks in the simulation. For example, in an episode of “stack two black blocks,” the system repeatedly tries to stack the blocks when the block falls. This suggests that the architecture is capable of learning recovery behavior. We show examples for three tasks: “stack blocks,” “place cups,” and “slide block to the color target.”, whose videos are on the project website.

\section{Conclusions and Limitations}
\label{sec:discussion}
In this work, we proposed \method, a fast and precise model for 3D object manipulation. It is built on the prior state-of-the-art RVT. Using a combination of architectural and system-level improvements, we significantly improved the speed, precision, and task performance. Although none of the techniques we used is novel in itself, our contribution lies in combining them effectively to advance the state-of-the-art in few-shot 3D manipulation. We found that \method significantly outperforms prior methods on RLBench while requiring much less compute. In the real world, we found that \method can solve high-precision tasks that involve inserting pegs and plugs using a single third-person camera and with just 10 demonstrations. 

We identify various limitations of \method which could be avenues of future work. \method, like RVT and PerAct, works with object instances that it was trained on. Extending this to unseen object instances would be an exciting direction. Although on high precision tasks, \method achieves surprising success with just a single RGB-D sensor, it sometimes fails due to minor insertion position errors. Augmenting \method to use force information to adjust fine-grained motions could be very interesting. As seen with the \textit{open drawer} task for \method, multi-task optimization could worsen performance on some tasks as training progresses. Developing a strategy to prevent this would be very useful. Lastly, although \method improves the overall performance on multi-task 3D manipulation by 17 points, the task is still far from being solved with \method achieving a success rate of 82\% in simulation and 72\% in the real world.

\smallsec{Acknowledgement} We thank Karl Van Wyk for fabricating our Franka's fingers.

\bibliographystyle{plainnat}
\bibliography{references}
\end{document}